\documentclass[10pt,twocolumn,letterpaper]{article}

\usepackage{cvpr}              
\definecolor{cvprblue}{rgb}{0.21,0.49,0.74}
\usepackage[pagebackref,breaklinks,colorlinks,allcolors=cvprblue]{hyperref}
\usepackage{xcolor}         
\usepackage{graphicx}
\usepackage{booktabs}
\usepackage{multirow}
\usepackage{multicol}
\usepackage{bm}
\usepackage{algorithm} %
\usepackage{amssymb}
\usepackage{amsmath}
\usepackage{algorithmic} %
\usepackage{amsmath}
\usepackage{color}
\usepackage{xcolor}
\usepackage{colortbl}
\usepackage{makecell}
\definecolor{lightgray}{gray}{0.9}

\def\paperID{12172} 
\def\confName{CVPR}
\def\confYear{2025}

\title{WeatherGen: A Unified Diverse Weather Generator for LiDAR Point Clouds via Spider Mamba Diffusion}

\author{Yang Wu$^1$ \quad  Yun Zhu$^{1}$ \quad Kaihua Zhang$^2$ \quad Jianjun Qian$^1$ \quad Jin Xie$^{3,4}$\thanks{Corresponding author.} \quad Jian Yang$^1$\\
$^1$PCA Lab, Nanjing University of Science and Technology, Nanjing, China\\
$^2$School of Automation, Southeast University, Nanjing, China\\
$^3$State Key Laboratory for Novel Software Technology, Nanjing University, Nanjing, China\\
$^4$School of Intelligence Science and Technology, Nanjing University, Suzhou, China\\
{\tt\small \{wuyang98, zhu.yun, csjqian, csjyang\}@njust.edu.cn; zhkhua@gmail.com; csjxie@nju.edu.cn}
}

\begin{document}
\maketitle
\begin{abstract}
  3D scene perception demands a large amount of adverse-weather LiDAR data, yet the cost of LiDAR data collection
  presents a significant scaling-up challenge.
  To this end, a series of LiDAR simulators have been proposed. Yet, they can only simulate a single adverse weather with a single physical model, and the fidelity of the generated data is quite limited.
  This paper presents \textbf{WeatherGen}, the first unified diverse-weather LiDAR data diffusion generation framework, significantly improving fidelity.
  Specifically, we first design a map-based data producer, which can provide a vast amount of high-quality diverse-weather data for training purposes.
  Then, we utilize the diffusion-denoising paradigm to construct a diffusion model. 
  Among them, we propose a spider mamba generator to restore the disturbed diverse weather data gradually.
  The spider mamba models the feature interactions by scanning the LiDAR beam circle or central ray, excellently maintaining the physical structure of the LiDAR data.
  Subsequently, following the generator to transfer real-world knowledge, we design a latent feature aligner.
  Afterward, we devise a contrastive learning-based controller, which equips weather control signals with compact semantic knowledge through language supervision, guiding the diffusion model to generate more discriminative data.
  %
  %
  Extensive evaluations demonstrate the high generation quality of WeatherGen. 
  Through WeatherGen, we construct the mini-weather dataset, promoting the performance of the downstream task under adverse weather conditions. Code is available: \url{https://github.com/wuyang98/weathergen}
\end{abstract}

\begin{figure}[!t]
\includegraphics[width=0.48\textwidth]{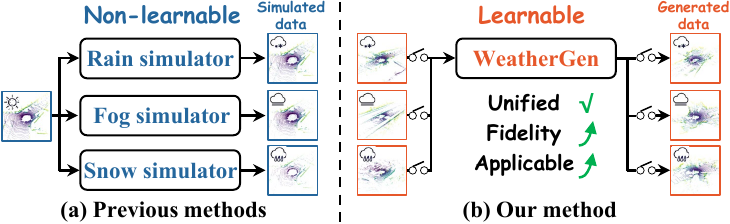}
\caption{
(a) Previous simulation-based methods~\cite{hahner2022lidar, hahner2021fog, kilic2021lidar, charron2018noising} can only provide a single non-learnable simulator for a single weather condition. 
Due to the complexity of optical propagation, previous simulated data all suffer from insufficient fidelity issues.
(b) Our method is the first unified generative framework. Through learning, the generated data has higher fidelity and can be more conducive to promoting downstream tasks under diverse weather.
}
\vspace{-0.4cm}
\label{fig: introduction}
\end{figure}
\vspace{-0.6cm}
\section{Introduction}
\label{sec: intro}
With the advent of large-scale autonomous driving datasets, 
a multitude of innovative solutions for 3D perception tasks, such as detection and trajectory planning emerge continuously.
Despite extensive research and satisfactory performance under clear weather, 3D scene perception under adverse weather is less studied due to the scarcity of large-scale LiDAR datasets in such scenarios~\cite{kong2023robo3d, bijelic2020seeing}.

Under adverse weather, lasers scatter and diffract, causing point noise and drop~\cite{park2024rethinking}. This leads to serious reliability issues, limiting the application expansions of existing models.
Unfortunately, collecting LiDAR data under adverse weather is very time, labor, and cost intensive~\cite{hahner2021fog, hahner2022lidar, bijelic2020seeing}.
Some simulation-based methods try to improve this issue through non-learnable simulators~\cite{hahner2022lidar, hahner2021fog, kilic2021lidar, charron2018noising}. 
As shown in Figure~\ref{fig: introduction}, they utilize a single physical model for a single condition. Due to the incomplete knowledge of sensor parameters and the complexity of optical propagation, these methods inevitably suffer from domain shifts and poor fidelity compared to real-world data.

Thus far, an inevitable question arises: \textit{can we construct a unified and high-fidelity diverse-weather LiDAR data generation framework?}
To achieve this goal, we have to solve three challenging points:
{(1) High-quality data Producer.}  LiDAR data are scarce under diverse weather conditions, which fails to meet the high-fidelity training requirements for generative models.
{(2)} High-fidelity Generator. The LiDAR data under diverse weather conditions possess noise and point drop characteristics, which present significant challenges to generative models.
{(3)} High-discriminability Controller. The generator lacks an effective way to obtain control signals with compact semantic knowledge, causing a mismatch when generating diverse LiDAR data in a unified generation framework.

To this end, we propose a unified LiDAR data generation framework, named WeatherGen, which can cost-effectively generate LiDAR data with high fidelity and high discriminability.
Firstly, we adopt a training strategy of pre-training plus fine-tuning. A map-based data producer (MDP) is designed to provide diverse weather LiDAR data for model pre-training.
Unlike previous simulators~\cite{hahner2022lidar, hahner2021fog, kilic2021lidar, charron2018noising}, which are non-learnable, MDP uses learnable parameters to align the pre-training data with the real-world distribution.
After pre-training, we fine-tune all parameters except for MDP to enhance model performance further.
Secondly, different from previous LiDAR data generation models~\cite{zyrianov2022learning, nakashima2023lidar, wu2024text2lidar, ran2024towards}, which disrupt the structural information of LiDAR data, we design a spider mamba generator (SMG) with the spider mamba scan that better adapts to the LiDAR imaging process. 
Unlike vision mamba~\cite{liu2024vmamba, zhu2024vision}, which models on image patches, spider mamba models on the beam circles and central rays of the LiDAR point cloud, effectively preserving the structural information of LiDAR data.
%
%
Following SMG, we further design a latent feature aligner (LFA) that uses real-world data to jointly constrain the predictions of the SMG, which is beneficial for our generator to produce more realistic data.
Thirdly, we design a contrastive learning-based controller (CLC) that utilizes stylized LiDAR data for controllable generation. 
Compared to other control signals, stylized LiDAR data has better feature consistency and integrability~\cite{wang2024disentangled}. 
CLC contains a contrastive learning mechanism with the language supervision from CLIP~\cite{radford2021learning}, helping the weather control signal learn more compact and discriminative semantic knowledge.
In summary, our contributions are listed as follows:
\begin{itemize}
\item We propose WeatherGen, the first unified framework for generating diverse weather LiDAR data. It contains a specially designed MDP to produce diverse weather LiDAR data in a map-to-map manner for WeatherGen training.
\item An SMG is devised, which can model the LiDAR feature interactions effectively in a way that can best maintain the physical structure of the LiDAR data. An LFA is followed to transfer real-world knowledge into the SMG.
\item A CLC is proposed to guide the model in generating more accurately, where contrastive learning is applied to equip the control signal with compact knowledge.

\item Extensive experiments demonstrate the superiority of WeatherGen in the generation quality of LiDAR data. Additionally, a diverse-weather LiDAR dataset is constructed through WeatherGen and applied to 3D object detection, revealing that generated data can promote the downstream task under adverse weather conditions.

\end{itemize}

\section{Related Work}
\label{sec: related work}
\textbf{LiDAR Data Simulation.}
%
%
Adverse weather conditions can lead to a noticeable decline in model performance, and in image tasks, certain methods~\cite{sakaridis2018model, sakaridis2018semantic, tremblay2021rain}alleviate this impact through simulation.
In LiDAR tasks, early method DROR~\cite{charron2018noising} proposes a 3D outlier detection algorithm to denoising data under adverse weather conditions.
Recently, Seeing Through Fog~\cite{bijelic2020seeing} contributes a small-scale diverse weather LiDAR dataset, providing a foundation for the research and evaluation of diverse weather LiDAR data. 
Since then, some simulation-based methods utilize the principles of optical propagation to simulate data for single weather conditions.
FSRL~\cite{hahner2021fog} proposes an optics-based method to simulate fog and uses simulated data to train a 3D object detection model, establishing a reliable method for data validity verification. 
LSS~\cite{hahner2022lidar} further considers the impact of wet ground and proposes a method to simulate snow. 
LISA~\cite{kilic2021lidar} designed simulation models with different parameters for rain, snow, and fog, all of which can assist in 3D object detection tasks under diverse weather conditions.
These methods can only design a single model for a single weather, and due to the complexity of optical propagation, the realism of the simulated data is not very good.

\noindent\textbf{LiDAR Data Generation.}
Compared to object-level and indoor point clouds, LiDAR point clouds are often more sparse and noisy~\cite{zhu2024spgroup3d, qu2024conditional, yan2024tri, hu2024rangeldm}, the common practice is to project LiDAR data onto a range map to obtain a more compact and regular form. 
This operation will not result in data loss~\cite{meyer2019lasernet, milioto2019rangenet}. On this basis, several studies are proposed.
Nakashima $et \ al.$~\cite{nakashima2021learning, nakashima2023generative} design a drop method to assist the training of GANs, achieving the generation of diverse sizes.
LiDARGen~\cite{zyrianov2022learning} can simulate the ray drop in LiDAR, and verify the feasibility of using diffusion models for LiDAR data generation.
UltraLiDAR~\cite{xiong2023learning} learns a compact 3D representation for LiDAR generation and completion.
The current work, LiDM~\cite{ran2024towards}, R2DM~\cite{nakashima2023lidar} and Text2LiDAR~\cite{wu2024text2lidar} design more mature frameworks that based on the denoising diffusion probabilistic model (DDPM), significantly improve performance. 
To date, there is no systematic study on the generation of diverse weather LiDAR data.

\begin{figure*}[!t]
\includegraphics[width=1\textwidth]{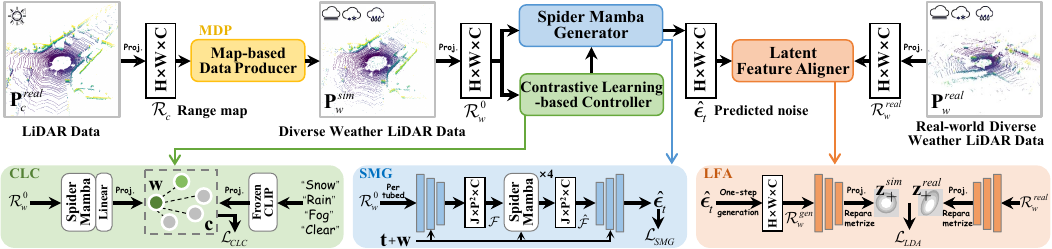}
\caption{The pipeline of WeatherGen. It has three core components. 
An MDP to produce high-quality training data that is closer to real-world data; 
An SMG models denoising features in a way that conforms to the LiDAR imaging process. SMG is further followed by an LFA to transfer real-world data knowledge into the generator.
A CLC, composed of a weather encoder and a CLIP text encoder, is used to generate control signals with more compact and discriminative knowledge through contrastive learning.
}
\label{fig: flowchart}
\end{figure*}

\noindent\textbf{State Space Models.}
The state space model (SSM) is widely adopted in various disciplines. Its core idea is connecting the input and output sequences using a latent state~\cite{xu2024survey}.
%
%
Structured SSM~\cite{gu2021efficiently} optimizes the computational cost of SSM, leading to a series of new SSM frameworks~\cite{gu2022parameterization, gupta2022diagonal, orvieto2023resurrecting, smith2022simplified}.
Mamba~\cite{gu2023mamba} further improves the SSM by integrating a selection mechanism, enabling it to selectively propagate or forget information, bringing a leap in performance.
%
%
%
%
Since then, a multitude of researchers attempt to transfer the success of mamba to the field of computer vision~\cite{liu2024vmamba, zhu2024vision, liang2024pointmamba}. 
%
%
Mamba also achieves notable success in the generative domain~\cite{ye2024p, oshima2024ssm, hu2024zigma}, benefiting from its excellent denoising capabilities. 
A robustness study~\cite{du2024understanding} demonstrates that the mamba possesses better robustness and generalization than the transformer.
Utilizing mamba for LiDAR data generation is worth exploring.

\section{Method}
We first introduce the basics of DDPM and SSM in Section~\ref{sec: ddpmssm}. Then, in Section~\ref{sec: data}, we describe the method for acquiring diverse weather LiDAR data for training. Afterward, in Section~\ref{sec: SMG}, we introduce the generator specifically designed for LiDAR data. Then, in Section~\ref{sec: CLC}, we introduce a method for obtaining discriminative control signals, and in Section~\ref{sec: loss}, we introduce the loss function.
\subsection{Preliminaries of DDPM and SSM}
\label{sec: ddpmssm}
We provide some foundational introductions to DDPM and SSM, and Figure~\ref{fig: flowchart} illustrates the training process of our method. After training is completed, we utilize only CLC and SMG for inference to generate data. 

The goal of DDPM is to learn a distribution, which operates in three parts~\cite{ho2020denoising}.
(1) Diffusion process: We continuously add noise until it approaches pure noise. The single-step noise addition process is:
$\mathcal{R}_{w}^{t} = \sqrt{\alpha_{t}}\mathcal{R}_{w}^{t-1} +  \sqrt{1 - \alpha_{t}}\bm{\epsilon}_{1}$,
where $\alpha_{t}=1-\beta_{t}$, $\beta_{t}$ is the noise weight, and the $\bm{\epsilon}_{1}$ is the added noise. Through iterative derivation~\cite{ho2020denoising}, we can obtain
$\mathcal{R}_{w}^{t} = \sqrt{\bar{\alpha}_{t}}\mathcal{R}_{w}^{0} +  \sqrt{1 - \bar{\alpha}_{t}}\bm{\epsilon}_{t}$,
where $\bar{\alpha}_{t}=\alpha_{t}\cdot \alpha_{t-1}...\alpha_{1}$, and $\bm{\epsilon}_{t}\sim \mathcal{N}(\textbf{0},\textbf{I})$. 
The aforementioned process can be further expressed as:
$q(\mathcal{R}_{w}^{t}|\mathcal{R}_{w}^{0})=\mathcal{N}(\mathcal{R}_{w}^{t};\sqrt{\bar{\alpha}_{t}}\mathcal{R}_{w}^{0}, (1-\bar{\alpha}_{t})\textbf{I})$.
(2) Inference process: This is the process of recovering the data from Gaussian noise. 
%
\begin{equation}
q(\mathcal{R}_{w}^{t-1}|\mathcal{R}_{w}^{t},\mathcal{R}_{w}^{0})=\mathcal{N}(\mathcal{R}_{w}^{t-1};\widetilde{\mu}_{t}(\mathcal{R}_{w}^{t},\mathcal{R}_{w}^{0}),\widetilde{\beta}_{t}\textbf{I}),
\label{eq: epsilon}
\end{equation}
in which, $\widetilde{\beta}_{t}=\frac{1-\bar{\alpha}_{t-1}}{1-\bar{\alpha}_{t}}\cdot\beta_{t}, 
\widetilde{\mu}_{t}(\mathcal{R}_{w}^{t}, \mathcal{R}_{w}^{0})=\frac{1}{\sqrt{\alpha_{t}}}(\mathcal{R}_{w}^{t}-\frac{\beta_{t}}{\sqrt{1-\bar{\alpha}_{t}}}\hat{\epsilon}_{t})$, and $\hat{\epsilon}_{t}$ is predicted by our SMG.
(3) Training process: The training objective is to use the denoising network to estimate the noise $\hat{\epsilon}_{t}$ in Equation~\ref{eq: epsilon}.

%
The continuous system inspires SSMs. Through the hidden state $h(s)\in \mathbb{R}^{N}$~\cite{zhu2024vision, gu2021efficiently}, SSMs can map a 1-D sequence $x(s)\in \mathbb{R} \rightarrow y(s)\in \mathbb{R}$. These models excel in representing systems
using a set of first-order differential equations, capturing the dynamics of the system’s state variables:
$h'(s) = \textbf{A}h(s) + \textbf{B}x(s), \quad
y(s) = \textbf{C}h(s)$
, where $h'(s)$ is the time derivative of the state vector $h(s)$. $\textbf{A}\in \mathbb{R}^{N\times N}$ is the evolution parameter, $\textbf{B}\in \mathbb{R}^{N\times 1}$ and $\textbf{C}\in \mathbb{R}^{1\times N}$ refer to the projection parameters that measure the relationships between $h(s)$, $x(s)$ and $y(s)$.
To convert it into a discrete form, a common practice is to utilize the zero-order holds, which are defined as: 
$
\overline{\textbf{A}} = \texttt{exp}(\bm{\Delta}\textbf{A}),\quad \overline{\textbf{B}}=(\bm{\Delta}\textbf{A})^{(-1)}(\texttt{exp}(\bm{\Delta}\textbf{A})-\textbf{I})\cdot \bm{\Delta}\textbf{B}$,
$\bm{\Delta}$ is the stepsize parameter, $\overline{\textbf{A}}$ and $\overline{\textbf{B}}$ are the discrete parameters. The discretized version using $\bm{\Delta}$ can be rewritten as:
\begin{equation}
h_k = \overline{\textbf{A}}h_{k-1} + \overline{\textbf{B}}x_k, \quad
y_k = \textbf{C}h_k.
\end{equation}
After these, we can obtain the output through a global convolution,
$\overline{\textbf{K}} = (\textbf{C}\overline{\textbf{B}}, \textbf{C}\overline{\textbf{AB}},...,\textbf{C}\overline{\textbf{A}}^{m-1}\overline{\textbf{B}}), \quad
\textbf{y} = \textbf{x} * \overline{\textbf{K}}
$,
where $m$ is the length of the input sequence $\textbf{x}$, and $\overline{\textbf{K}}$ is a structured convolution kernel with size $m$.

\subsection{Map-based Data Producer}
\label{sec: data}
In this part, we aim to obtain the diverse weather LiDAR data through MDP for training. 
Under adverse weather, point cloud often exhibits additional noise points or point drops~\cite{park2024rethinking}. 
Different weather conditions have their customizations.
\noindent\textbf{Snow:}
Snowflakes will cause irregular reflections, leading to the appearance of random noise points and the loss of some points~\cite{seppanen20224denoisenet}.
\noindent\textbf{Fog:}
The liquid in fog is fine and dense, causing all lasers to be attenuated, which is particularly noticeable at long distances. Lasers also generate a lot of noise due to irregular refraction in the fog~\cite{hahner2021fog}.
\noindent\textbf{Rain:}
Water droplets also can result in fewer noise points. Due to the ground becoming wet, points at a distance may be lost due to large angle reflections~\cite{ kilic2021lidar}.
These result in an exceptionally complex process of laser propagation, making it difficult to model exhaustively with fixed parameters.

%
%
Unlike previous non-learnable simulators~\cite{hahner2022lidar, hahner2021fog, kilic2021lidar, charron2018noising}, we additionally provide learnable masks, endowing MDP with the ability to learn from data~\cite{park2024rethinking}.
%
Specifically, given LiDAR data $\textbf{P}_{c}^{real} \in \mathbb{R}^{4}$ with coordinates and intensity $(p^{x},p^{y},p^{z},p^{i})$ under clear weather, we first convert it into a range map $\mathcal{R}_{c}$. Through a map-to-map approach, we can produce the diverse-weather data:
\begin{equation}
\begin{split}
&\mathcal{R}_{w} = \texttt{MDP}(\mathcal{R}_{c}) \\
&= \left\{
\begin{aligned}{}
& \mathcal{R}_{c} \odot \texttt{BDF}(\mathcal{M}_{e1}, r_{w})+\mathcal{R}_{n}+\mathcal{M}_{d}, d>r_{w}, \\
& \mathcal{R}_{c} \odot \texttt{BDF}(\mathcal{M}_{e2}, r_{w})+\mathcal{R}_{n} + \mathcal{M}_{d}, d<r_{w}, 
\end{aligned}
\right.
\end{split}
\label{eq: ls}
\end{equation}
where $w\in [snow, fog, rain]$, $\mathcal{M}_{e}$ is the mask composed of $0$ and $1$ to simulate the attenuation of the laser at a distance. 
When the point satisfies $d>r_{w}$, the corresponding position is set to 0, otherwise, it is set to 1, and $d$ is the distance value, which can be obtained from $\textbf{P}$.
This part ensures that the produced data has the most basic diverse-weather characteristics, leading to a reliable initialization.
$\mathcal{M}_{d}$ is the mask learned through Equation~\ref{eq: finalloss} to adaptively adjust parameters. 
This part ensures our MDP can dynamically update and adaptively align with real-world distribution.
$\mathcal{R}_{n}$ is the random noise related to the $r_{w}$.
$r_{w}$ is an adjustable parameter that reflects the severity of adverse weather, the smaller the value, the more adverse it represents, and more settings can be found in the Appendix.
$\texttt{BDF}$ is a Bernoulli distribution function~\cite{ristic2013tutorial} for random dropping, which can simulate the noisier state of LiDAR data in adverse weather, the drop rate is related to $r_{w}$.

\begin{figure}[!t]
\includegraphics[width=0.48\textwidth]{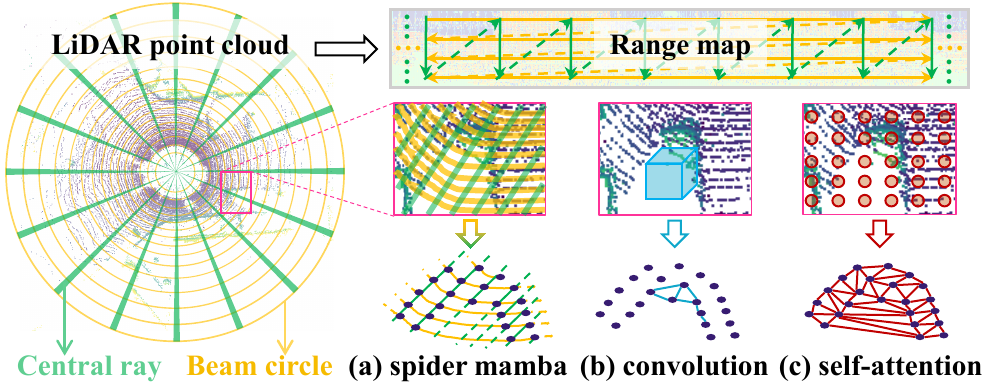}
\caption{(a) Spider mamba scans model features along the LiDAR beams-circle and central ray, excellently maintaining the physical structures of the LiDAR; (b) Convolution can only model features locally; (c) Self-attention disorderly connecting all points, disrupting the physical properties of LiDAR data.}
\vspace{-0.35cm}
\label{fig: mamba}
\end{figure}

\subsection{Spider Mamba Generator}
\label{sec: SMG}
As stated in Section~\ref{sec: related work}, the range map is proven to be a successful form for processing LiDAR data.
However, previous methods~\cite{zyrianov2022learning, nakashima2023lidar, ran2024towards, wu2024text2lidar} utilizing convolution and self-attention on the range map, do not adequately accommodate the annular structure of LiDAR data. 
In contrast, in SMG, we design a mamba-based method to model along the LiDAR beam circles or central rays, which corresponds exactly to the rows or columns of the range map and is more consistent with the imaging method of LiDAR. As shown in Figure~\ref{fig: mamba}, this progress is akin to a spider hunting on a web, hence we call it “spider mamba”.

Unlike the successful mamba methods in image tasks~\cite{liu2024vmamba, zhu2024vision, hu2024zigma} that model features on patches, our spider mamba makes more suitable designs tailored for LiDAR data.
We follow the practice in UltraLight VM-UNet~\cite{wu2024ultralight}, and no longer perform scanning at the patch level but at the point level.
Rows and columns of the range map correspond to the beam circles and central rays of the LiDAR point cloud. 
Patches of range maps correspond to a cluster of point clouds, which, in sparse outdoor scenarios, can lose the geometric significance of LiDAR data.
Figure~\ref{fig: mamba} also presents a comparison between spider mamba, convolution (Conv) and self-attention (SA).
%
%
%
Spider mamba models along the LiDAR's beam circle and central ray, while Conv models locally and SA connects all points, disrupting the physical properties of LiDAR and diluting a significant amount of beam circle information. 

As shown in Figure~\ref{fig: flowchart}, the SMG we devised adopts a U-shape denoising structure, with both the decoder and encoder consisting of convolutional blocks and spider mamba blocks. A more detailed visual structure diagram can be found in the Appendix.
Specifically, we first use convolution to map $\mathcal{R}_{w}^{t}$ to latent features and further expand them into pixel sequences to obtain $\mathcal{F}\in \mathbb{R}^{J\times C}$, where $J$ is the number of pixels.
%
%
Following UltraLight VM-UNet~\cite{wu2024ultralight}, to save computational cost, we decompose $\mathcal{F}$ into $\mathcal{F}_{i} \in \mathbb{R}^{J\times (C/4)},i\in(1,4)$ by channels.
Then, we linearly project each $\mathcal{F}_{i}$ to the embedding with size $D$ as:
\begin{equation}
\textbf{T}_{i}^{0} = [\mathcal{F}^{1}_{i}\textbf{W}; \mathcal{F}^{2}_{i}\textbf{W},...,\mathcal{F}^{j}_{i}\textbf{W}],
\end{equation}
where $\mathcal{F}^{j}_{i}$ is the $j$-th patch of $\mathcal{F}_i$, $\textbf{W}\in \mathbb{R}^{(C/4)\times D}$ is the learnable projection matrix. 
Then, we send the sequence $\textbf{T}_{i}^{l-1}$ into the $l$-th layer of the mamba, getting the output $\textbf{T}_{i}^{l}$. 
For a more detailed step-by-step mamba process, see vision mamba~\cite{zhu2024vision}.
%
%
Once we obtain the output sequence $\textbf{T}^{L}_{i}$, we normalize it and feed it into the multi-layer perceptron
(MLP) head to get the final prediction $\hat{\mathcal{F}}_{i}$. Each $\hat{\mathcal{F}}_{i}$ is concatenated to form the final latent features $\hat{\mathcal{F}}$. 
%
%
%
After passing through the decoder of the denoising network, 
we can get the predicted noise $\hat{\bm{\epsilon}}_{t}$ at a corresponding timestep. 
We use the mean squared error loss function to supervise the learning process:
\begin{equation}
\begin{split}
&\mathcal{L}_{SMG} = \mathcal{L}_{MSE}(\bm{\epsilon}_{t}, \hat{\bm{\epsilon}}_{t})
\\
&= \mathbb{E}_{t \sim U(0, T),\bm{\epsilon} \sim \mathcal{N}(0,1)} [||\bm{\epsilon}_{t} - \texttt{SMG}(\mathcal{R}_{w}^{t},\textbf{t},\textbf{w};\bm{\theta})||^{2}],
\label{eq: SMGloss}
\end{split}
\end{equation}
where $\bm{\theta}$ is the learnable parameters of the SMG, $\textbf{t}$ is the timestep embedding, $\textbf{w}$ is the control signal from the CLC. 

\noindent\textbf{Latent Feature Aligner.}
In order to leverage real-world data to drive the model to learn a more realistic data distribution, we further designed the LFA as a subsequent step following SMG to transfer real-world knowledge.
%
%
We utilize the predicted noise $\hat{\bm{\epsilon}}_{t}$ for one-step generation $\mathcal{R}_{w}^{gen}=\frac{1}{\sqrt{\bar{\alpha}_{t}}}(\mathcal{R}_{w}^{t}-\sqrt{1-\bar{\alpha}}\hat{\epsilon}_{t})$. 
One-step generation may not fully recover the range map, but its quality is sufficient to provide weather feature information~\cite{liu2023instaflow, song2023consistency}.

Then, we introduce the real-world diverse LiDAR data $\mathcal{R}_{w}^{real}$ and project both $\mathcal{R}_{w}^{real}$ and $\mathcal{R}_{w}^{gen}$ into a unified latent space $\mathcal{N}(\bm{\mu}, \bm{\sigma})$ through linear projection to model their weather patterns. We randomly sample the weather latent variable $\textbf{z}^{s}\sim \mathcal{N}(\bm{\mu}^{s}, \bm{\sigma}^{s}), s\in [gen, real]$ and use the Kullback-Leibler Divergence to align them:
\begin{equation}
\small
\begin{split}
&\mathcal{L}_{LFA} = \mathcal{L}_{KL}(P_{real}(\textbf{z}^{real}|\mathcal{R}_{w}^{real})||Q_{gen}(\textbf{z}^{gen}|\mathcal{R}_{w}^{gen})) \\
&= \sum P_{real}(\textbf{z}^{real}|\mathcal{R}_{w}^{real})\mathtt{log}\frac{P_{real}(\textbf{z}^{real}|\mathcal{R}_{w}^{real})}{Q_{gen}(\textbf{z}^{gen}|\mathcal{R}_{w}^{gen})}.
\label{eq: DAloss}
\end{split}
\end{equation}
$\textbf{z}^{s}$ is reparameterized from a pair of $(\bm{\mu}^{s}, \bm{\sigma}^{s})$:
$\textbf{z}^{s} = \bm{\sigma}^{s} \odot \bm{\eta} + \bm{\mu}^{s}$, and $\bm{\eta}\sim\mathcal{N}(\textbf{0},\textbf{I})$ is the random noise.
Each position in the latent space represents a potential diverse data that could expand the boundaries of weather distributions from the small-scale real-world data~\cite{wu2023co,zhang2021uncertainty}, and enhance the effectiveness of transferring real-world knowledge.


\subsection{Contrastive Learning-based Controller}
\label{sec: CLC}
To achieve a unified diverse weather generation framework, learning the control signal with compact semantic knowledge is also important, which can guide the model in discriminatively generating under diverse weather conditions.

For this, we design a contrastive learning-based controller, which includes a weather encoder $\mathcal{W}$ and a frozen CLIP text-encoder~\cite{radford2021learning} $\mathcal{C}$. The weather embedding $\textbf{w}$ and the semantic embedding $\bm{c}$ is obtained through:
\begin{equation}
\textbf{w}= \mathcal{W}(\mathcal{R}_{w}^{0}), \bm{c}_{i}= \mathcal{C}(``\small \texttt{Weather text prompts}").
\end{equation}
We pre-set four weather prompts and input them into $\mathcal{C}$ to obtain the corresponding text embedding $\bm{c}_{i},i=1,...,4$ for knowledge anchoring. Here, $\bm{c}_{j}$ corresponds to the embedding that is consistent with the weather state of $\mathcal{R}_{w}^{0}$. 
%
%
%
%
This process is essentially an information bottleneck process that utilizes the concept of contrastive learning~\cite{hu2024survey}:
\begin{equation}
\mathcal{L}_{CLC} = \texttt{IB}(\bm{\varphi})= I(\textbf{w},\textbf{c}_{i\neq j};\bm{\varphi}) - \beta \cdot I(\textbf{w},\textbf{c}_{i=j};\bm{\varphi}),
\label{eq: CLCloss}
\end{equation}
where $I$ represents mutual information, $\textbf{w}$ represents the compressed weather embedding, $\textbf{c}$ represents the target embedding with high discriminative power, and $\beta$ is the weighting parameter, empirically set to 0.2, $\bm{\varphi}$ is the learnable parameter of CLC.

Our optimization goal is to minimize $I(\textbf{w},\textbf{c}_{i\neq j};\bm{\varphi})$, to reduce the redundancy of $\textbf{w}$ to unrelated weather embeddings $\textbf{c}_{i\neq j}$ as much as possible, ensuring that the most critical information related to $\mathcal{R}_{w}$ is retained. At the same time, maximizing $I(\textbf{w},\textbf{c}_{i=j};\bm{\varphi})$ to help $\textbf{w}$ achieve a more compact semantic knowledge, which can guide the diffusion network to generate more information discriminative data. 
The $\textbf{w}$ will be concatenated with timestep $\textbf{t}$ and fused with the features in the SMG along the channel dimension.

\subsection{Loss Function}
\label{sec: loss}
\vspace{-0.1cm}
The final loss function used to pre-train our model is:
\begin{equation}
\vspace{-0.1cm}
\mathcal{L} = \mathcal{L}_{SMG} + \mathcal{L}_{LFA} + \mathcal{L}_{CLC},
\label{eq: finalloss}
\vspace{-0.1cm}
\end{equation}
where $\mathcal{L}_{SMG}$, $\mathcal{L}_{LFA}$ and $\mathcal{L}_{CLC}$ are sourced from Equation~\ref{eq: SMGloss}, Equation~\ref{eq: DAloss} and Equation~\ref{eq: CLCloss}, respectively.
After the pre-training is completed, we use real-world data and the same loss function to fine-tune (FT) the model, updating all parameters except for MDP. After the entire training process is completed, we utilize Equation~\ref{eq: epsilon} to generate data.

\begin{figure*}[!t]
\centering
\includegraphics[width=1\textwidth]{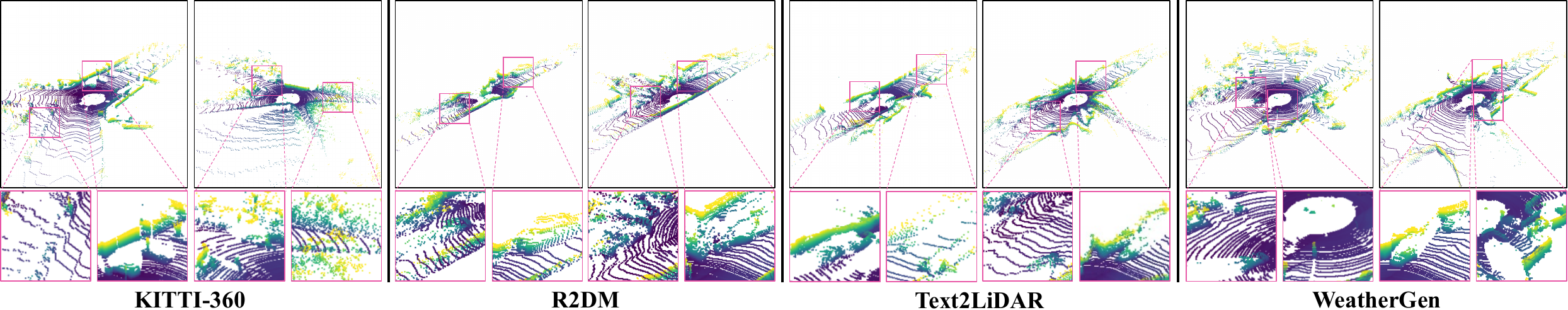}
\vspace{-0.4cm}
\caption{Visual comparisons with competitive generation methods on KITTI-360~\cite{liao2022kitti} dataset. The results prove that WeatherGen has a better ability to generate clear object contours and small objects.
}
\label{fig: visual}
\end{figure*}
%
\section{Experiments}
\subsection{Datasets and Evaluation Metrics}
We evaluate our model on KITTI-360~\cite{liao2022kitti}, a dataset primarily contains clear weather, and Seeing Through Fog~\cite{bijelic2020seeing} that contains snow, rain, and fog weather conditions. 
KITTI-360~\cite{liao2022kitti} has 50,348 frames for training and validation, and 26,367 frames for testing~\cite{zyrianov2022learning, nakashima2023lidar}. 
The adverse weather data size of Seeing Through Fog (about 200 to 3,000 frames for each weather)~\cite{bijelic2020seeing} is not enough to support training, we split it for fine-tuning and testing. 
WeatherGen can accommodate both unconditional and conditional generations. \textit{For unconditional generation}, the MDP and CLC are removed.
Following LiDARGen~\cite{zyrianov2022learning}, R2DM~\cite{nakashima2023lidar} and Text2LiDAR~\cite{wu2024text2lidar}, we calculate the distributional dissimilarity between 10,000 generated and real-world samples in three data formats: range map, point clouds, and bird's eye view (BEV).
Range map form utilizes the pre-trained RangeNet~\cite{milioto2019rangenet} to calculate the Frechet range map distance (FRD)~\cite{zyrianov2022learning}. 
Point cloud form uses the pre-trained PointNet~\cite{qi2017pointnet} to obtain the Frechet point cloud distance (FPD). 
BEV form uses Jensen–Shannon divergence (JSD) and minimum matching distance (MMD) to measure the distance between the marginal distributions of BEV occupancy grids.
\textit{For weather-conditional generation}, we calculate the distributional dissimilarity between 200 generated and real-world samples from different test splits of Seeing Through Fog. The metrics used are the same as \textit{unconditional generation}. 
WeatherGen can also perform the LiDAR point cloud densification task, more experimental details can be found in the Appendix.
\subsection{Unconditional LiDAR Data Generation}
\begin{table}[!t]
\huge
\centering
\renewcommand\arraystretch{1.15}
\caption{Quantitative unconditional generation comparisons with other state-of-the-art methods on KITTI-360~\cite{liao2022kitti} dataset. }
\vspace{-0.1cm}
\resizebox{0.48\textwidth}{!}{
\begin{tabular}{ccccccc}
\bottomrule
\multirow{2}{*}{Method}  & Point cloud & Range map  & \multicolumn{2}{c}{BEV occupancy grid} \\ \cmidrule(r){2-2}\cmidrule(r){3-3}\cmidrule(r){4-4}\cmidrule(r){5-5}\cmidrule{6-7} 
& FPD$\downarrow$ & FRD$\downarrow$ & MMD$\times10^{-4}\downarrow$ & JSD$\times10^{-2}\downarrow$ \\ \hline
LiDARGAN~\cite{caccia2019deep} & -  & 3003.8 & 30.60 & - \\
LiDARVAE~\cite{caccia2019deep}  & -  & 2261.5 & 10.00 & 16.10 \\
ProjectedGAN~\cite{sauer2021projected} & -  & 2117.2 & 3.47 & 8.50 \\
LiDARGen~\cite{zyrianov2022learning}  & 90.29 & 579.39 & 7.39 & 7.38 \\
LiDM~\cite{ran2024towards}   & 34.36 & 334.55 & 1.07 & 4.77 \\
R2DM~\cite{nakashima2023lidar}    & 6.24 & \underline{149.66} & 1.91 & 3.05 \\
Text2LiDAR~\cite{wu2024text2lidar}    & \textbf{4.81} & 164.16 & \underline{0.49} & \underline{2.01} \\
\rowcolor[gray]{0.9}
WeatherGen   & \underline{6.15} & \textbf{138.62} & \textbf{0.39} & \textbf{1.99} \\ 
\bottomrule
\vspace{-0.9cm}
\label{tab: unconditional_metrics}
\end{tabular}}
\end{table}
\begin{table}[!t]
\huge
\centering
\renewcommand\arraystretch{1.15}
\caption{
Quantitative comparisons with other state-of-the-art simulation methods on Seeing Through Fog~\cite{bijelic2020seeing} dataset.
}
\vspace{-0.1cm}
\resizebox{0.48\textwidth}{!}{
\begin{tabular}{ccccccc}
\bottomrule
\multirow{2}{*}{\makecell[c]{Method\\  \& weather}}  & Point cloud & Range map  & \multicolumn{2}{c}{BEV occupancy grid} \\ \cmidrule(r){2-2}\cmidrule(r){3-3}\cmidrule(r){4-4}\cmidrule(r){5-5}\cmidrule{6-7} 
& FPD$\downarrow$ & FRD$\times10^{1}\downarrow$ & MMD$\times10^{-4}\downarrow$ & JSD$\times10^{-1}\downarrow$ \\ \hline
LSS(Snow)~\cite{hahner2022lidar} & 106.37 & 142.17 & 3.59 & 2.11 \\
FSRL(Fog)~\cite{hahner2021fog} & 319.32 & 210.51 & 8.56 & 3.69 \\
LISA(Rain)~\cite{kilic2021lidar} & 301.11 & 145.23 & 4.30 & 3.23 \\
\rowcolor[gray]{0.9}
WeatherGen(Snow)   & \textbf{59.28} & \textbf{124.17} & \textbf{1.71} & \textbf{0.77} \\ 
\rowcolor[gray]{0.9}
WeatherGen(Fog)   & \textbf{314.14} & \textbf{196.89} & \textbf{8.08} & \textbf{2.66} \\
\rowcolor[gray]{0.9}
WeatherGen(Rain)   & \textbf{86.40} & \textbf{127.06} & \textbf{4.15} & \textbf{0.93} \\ \bottomrule
\vspace{-2cm}
\label{tab: weather_metrics}
\end{tabular}}
\end{table}
By removing the weather control signal $\textbf{w}$, WeatherGen can operate as an unconditional generative model, specifically generating data under clear weather conditions. 
We compare our approach with GAN~\cite{caccia2019deep, sauer2021projected}, VAE~\cite{caccia2019deep}, and diffusion~\cite{zyrianov2022learning, nakashima2023lidar, wu2024text2lidar, ran2024towards} based methods. 
Among them, LiDM~\cite{ran2024towards} cannot generate point cloud intensity, when testing its FRD, we use depth as a substitute for intensity.
As shown in Table~\ref{tab: unconditional_metrics}, WeatherGen achieves impressive performance. Compared to the bulky transformer structure of Text2LiDAR, WeatherGen has a faster speed and fewer parameters which can be found in Figure~\ref{fig: para}.

Figure~\ref{fig: visual} shows the visual comparison with real-world data~\cite{liao2022kitti}, R2DM and Text2LiDAR. 
It can be found that the objects generated by R2DM and Text2LiDAR are much noisier, and a large number of point clouds penetrate them.
In contrast, the generations of WeatherGen are more consistent with the physical characteristics of LiDAR. For example, the objects have clearer outlines and more complete content. 
%
These facts prove that our model can generate LiDAR data with a more realistic distribution and better diversity. More visualizations of unconditional generations and denoising steps analysis can be found in the Appendix.

\begin{figure*}[!t]
\includegraphics[width=1\textwidth]{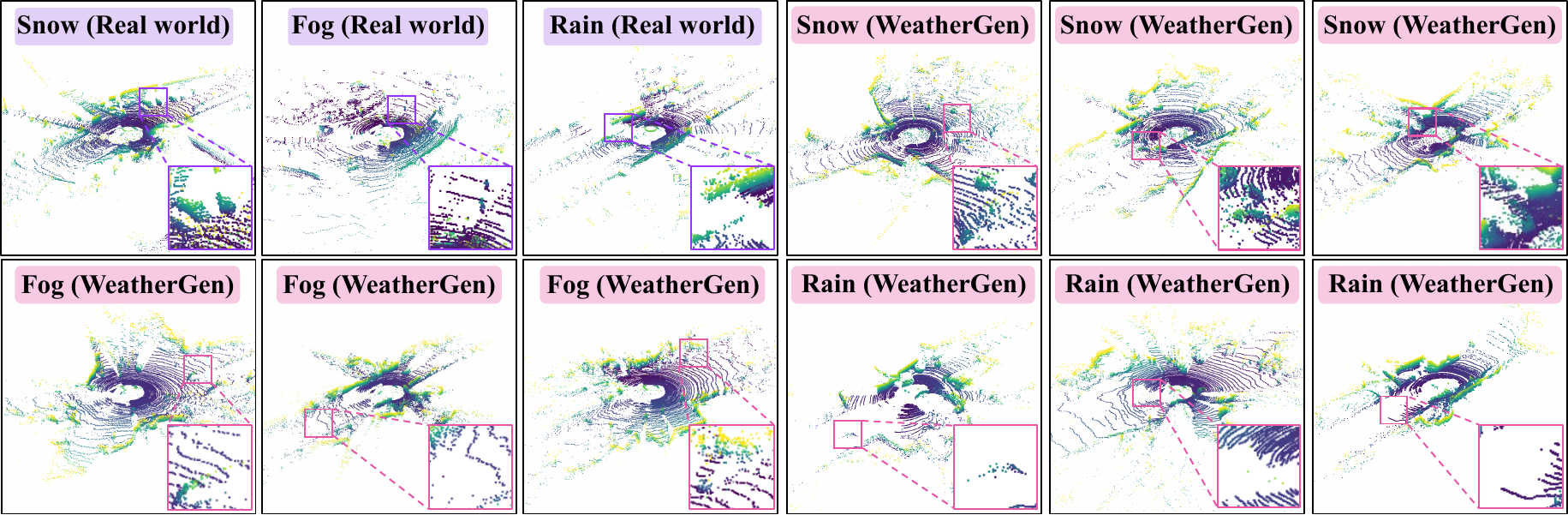}
\vspace{-0.6cm}
\caption{Visual comparisons of real-world diverse weather LiDAR data and generated results on Seeing Through Fog~\cite{bijelic2020seeing} dataset.}
\vspace{-0.4cm}
\label{fig: weather}
\end{figure*}
\vspace{-0.08cm}
\subsection{Weather-conditioned LiDAR Data Generation}
\vspace{-0.08cm}
Generating LiDAR point clouds under diverse weather conditions through a unified framework is the core,  distinctive, and pioneering feature of WeatherGen. 
On the Seeing Through Fog dataset~\cite{bijelic2020seeing}, we conduct tests using different splits: dense fog, heavy snow, and rain, these are the most representative weather data~\cite{bijelic2020seeing}.

Table~\ref{tab: weather_metrics} lists the quantitative comparison between WeatherGen and popular simulation-based methods~\cite{hahner2022lidar, hahner2021fog, kilic2021lidar}. 
The data generated by WeatherGen significantly outperforms previous simulation methods. 
This is because, compared to non-learnable simulators, WeatherGen can learn more complex optical propagation processes from real-world data, enhancing the fidelity of the generated data.

Figure~\ref{fig: weather} shows the comparison of generated and real-world LiDAR data. 
We can observe that the generated data closely aligns with real-world data in terms of imaging characteristics and weather properties.
More analysis of customized generation can be found in the Appendix. Section~\ref{sec: prompt} further introduces the value of our generations.

\subsection{Ablation study}
\begin{figure}[!t]\hspace{0.8mm}
\includegraphics[width=0.47\textwidth]{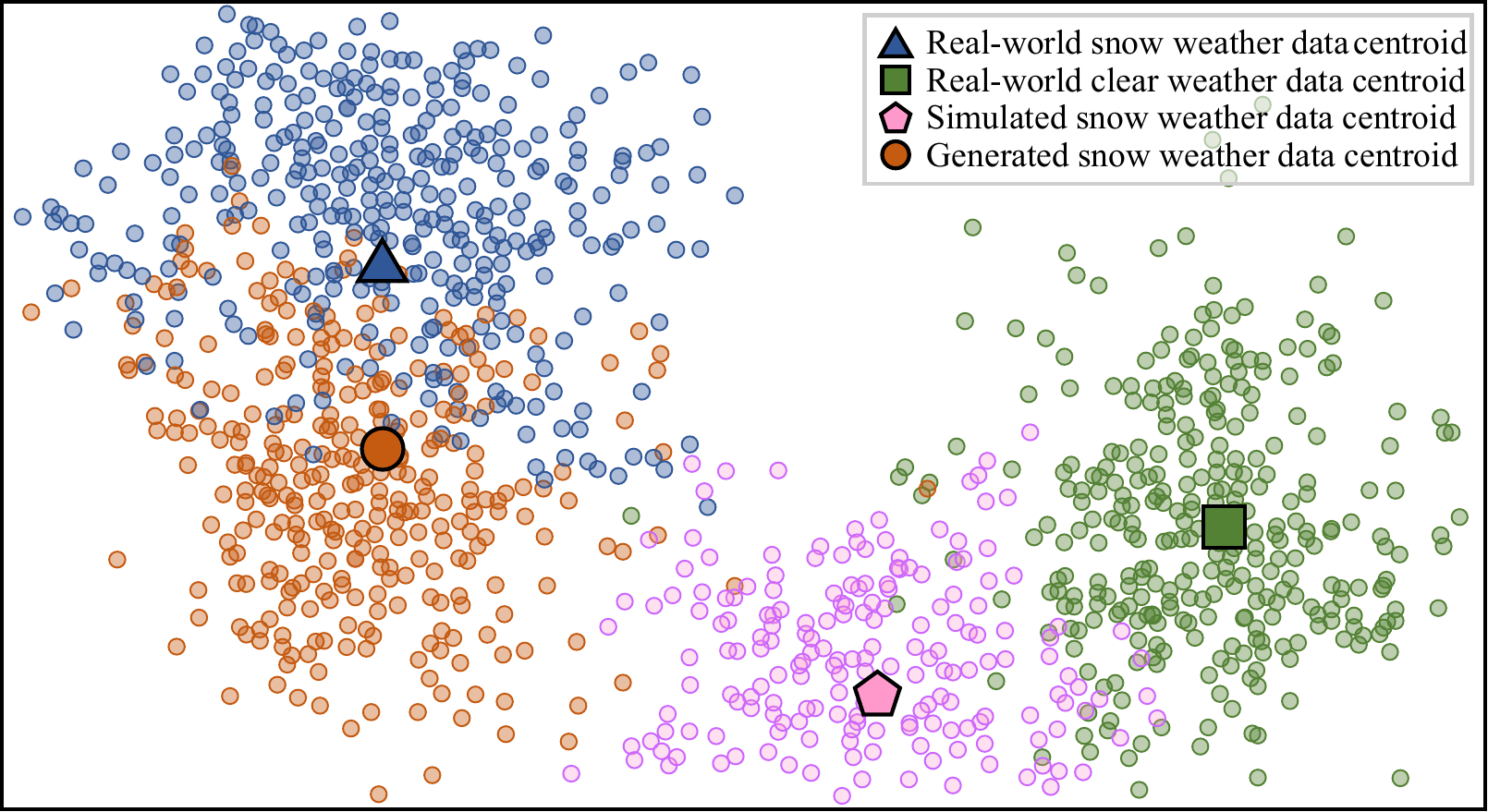}
\vspace{-0.5cm}
\caption{The t-SNE plots of the real-world data~\cite{bijelic2020seeing}, simulated snow weather data through MDP without $\mathcal{M}_{d}$ and FT, and generated snow weather data. Compared to simulated data, our generated data is better aligned with the real-world distribution.}
\label{fig: domain}
\end{figure}

We conduct ablation studies on the ``snow heavy'' test set of Seeing Through Fog.
When MDP is not used, it signifies that the learnable $\mathcal{M}_{d}$ is not employed.
When SMG is not used, we replace it with a U-Net~\cite{nakashima2023lidar}.
If CLC is not employed, only the weather encoder $\mathcal{W}$ is deployed. 
From Table~\ref{tab: ablation} we observe that when MDP is not used, a decline in performance occurs. 
Under adverse weather conditions, introducing learnable simulation parameters can better transfer real-world knowledge, enhancing the quality of generation.
Figure~\ref{fig: domain} further illustrates that when real-world data is not utilized at all, the generated data exhibits a greater deviation.
When SMG is not used, there is a performance decrease of varying degrees. 
This is because the LiDAR data in snowy scenes has more noise and experiences point drop.
The network is greatly disturbed and cannot well maintain the physical properties of the LiDAR lines without our SMG design.
When CLC is not used, the quality of the control signal decreases, resulting in a mismatch between the target weather and the generated weather, causing all metrics to decline.
LFA is also helpful for our model. WeatherGen can leverage LFA to directly align with the distribution of real-world data, which is key information that simulation-based methods cannot achieve at all. 
The use of "*-" represents training with small-scale diverse weather data from Seeing Through Fog, and due to the insufficiency of data volume, the performance is not optimal. 
These results demonstrate the effectiveness of our pre-training plus fine-tuning strategy, as well as the individual designs.
%

\subsection{Promoting 3D Object Detection}
\label{sec: prompt}
\begin{figure}[!t]
\includegraphics[width=0.48\textwidth]{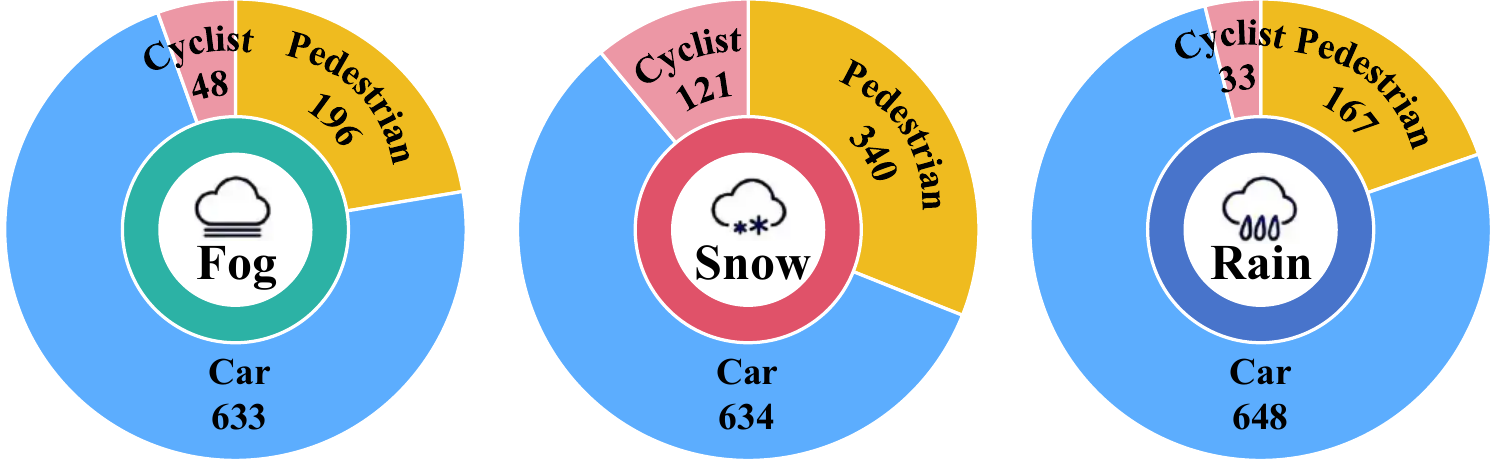}
\caption{The composition of the constructed mini-weather dataset. Each type of weather includes 256 frames, and we show the frequency of different objects appearing within them.}
\vspace{-0.4cm}
\label{fig: wea}
\end{figure}

To validate the practical value of the generated data, 
we first construct a mini-weather dataset using the generated data, providing 256 frames of data for each type of weather, and utilize LabelCloud~\cite{sager2021labelcloud} to label the objects within. Figure~\ref{fig: wea} illustrates the detailed composition of the mini-weather.

Then, we train the 3D object detection model using the mini-weather and compare the results with simulation-based data augmentation methods~\cite{charron2018noising, kilic2021lidar, hahner2021fog, hahner2022lidar}. 
Among them, DROR~\cite{charron2018noising}, FSRL~\cite{hahner2021fog}, and LSS~\cite{hahner2022lidar} can provide data augmentation for snowy and foggy conditions, while LISA~\cite{kilic2021lidar} can provide data augmentation for rainy, snowy, and foggy conditions.
To ensure fairness, we use the classic PointPillars~\cite{lang2019pointpillars} as the detector, train it on the ``train clear" train split from Seeing Through Fog~\cite{bijelic2020seeing} and test it on the ``dense fog", ``heavy snowfall," and ``rain" test splits.
When using the mini-weather, we replace the random 256 frames in train clear (3,469 frames) with the corresponding weather data from the mini-weather. This means that only $7.4\%$ of the training data comes from mini-weather.
Figure~\ref{tab: dense_fog},\ref{tab: heavy_snow},\ref{tab: rain_all} show the data validity test results under three weather conditions. Among the 21 metrics, the data generated by our method achieves the best results in 18.
From the last row of the table, it can be observed that using only $7.4\%$ of our data, compared to not using our data at all, can significantly enhance the performance of the detector. 

\begin{table}[!t]
\Huge
\centering
\renewcommand\arraystretch{1}
\caption{Ablation studies on Seeing Through Fog~\cite{bijelic2020seeing} dataset under snow heavy test split.}
\resizebox{0.48\textwidth}{!}{
\begin{tabular}{ccccc|cccc}
\bottomrule
\multicolumn{5}{c|}{Key designs} & Point cloud  & Range map &\multicolumn{2}{c}{BEV occupancy grid} \\  \cmidrule(r){6-6}\cmidrule(r){7-7}\cmidrule{8-9}
MDP & SMG & CLC & LFA & FT & FPD$\downarrow$ & FRD$\downarrow$ & MMD$\times10^{-4}\downarrow$ & JSD$\times10^{-1}\downarrow$  \\ \hline
$\checkmark$ & $\checkmark$ & $\checkmark$ & - & - & 244.61 & 1922.07 & 8.99 & 1.55 \\
$\checkmark$ & $\checkmark$ & $\checkmark$ & $\checkmark$ & - & 221.73 & 1887.51 & 7.85 & 1.59 \\
- & $\checkmark$ & $\checkmark$ & $\checkmark$ & - & 238.16 & 1945.36 & 9.22 & 1.78 \\
$\checkmark$ & $\checkmark$ & $\checkmark$ & - & -* & 148.25 & 1747.14 & 4.93 & 1.05 \\
$\checkmark$ & $\checkmark$ & - & $\checkmark$ & $\checkmark$ & 66.35 & 1423.69 & 2.28 & 0.94 \\
$\checkmark$ & - & $\checkmark$ & $\checkmark$ & $\checkmark$ & 79.45 & 1436.89 & 3.09 & 1.13 \\
- & $\checkmark$ & $\checkmark$ & $\checkmark$ & $\checkmark$ & 84.58 & 1501.79 & 3.41 & 1.85 \\
$\checkmark$ & $\checkmark$ & $\checkmark$ & - & $\checkmark$ & 77.39 & 1300.13 & 1.68 & 1.31 \\
\rowcolor[gray]{0.9}
$\checkmark$ & $\checkmark$ & $\checkmark$ & $\checkmark$ & $\checkmark$ & 59.28 & 1241.66 & 1.71 & 0.77 \\
        \bottomrule
\end{tabular}}
\label{tab: ablation}
\end{table}

It is worth mentioning that the time required to collect and create a thousand-frame level LiDAR dataset for adverse weather conditions is measured in years~\cite{bijelic2020seeing}. 
In contrast, the time needed to construct the mini-weather LiDAR dataset is just a few days. This indicates that WeatherGen has great value in enhancing the safety of unmanned systems under adverse weather conditions.
\begin{table}[!t]
\Huge
\centering
\renewcommand\arraystretch{1.15}
\caption{
Comparison with simulation-based methods on Seeing Through Fog~\cite{bijelic2020seeing} dataset under dense fog test split. 
}
\vspace{-0.1cm}
\resizebox{0.48\textwidth}{!}{
\begin{tabular}{ccccc|ccc}
\bottomrule
\multirow{2}{*}{Method}  & \multicolumn{7}{c}{Dense fog scenarios$\uparrow$}  \\ 
\cmidrule(r){2-8}
& 0-80m & 0-30m & 30-50m & 50-80m & \makecell[c]{Car AP\\ @.5IoU} & \makecell[c]{Cyc. AP\\ @.25IoU} & \makecell[c]{Ped. AP\\ @.25IoU} \\ \hline
DROR~\cite{charron2018noising} & 12.25 & 24.52 & 9.60 & 1.22 & 33.42 & \underline{23.89} & 30.73 \\
LISA~\cite{kilic2021lidar} & 15.66 & 33.01 & 8.96 & \underline{1.90} & 38.21 & 19.75 & \underline{31.39} \\
FSRL~\cite{hahner2021fog} & 14.56 & 31.85 & 7.89 & 1.34 & 35.94 & 21.20 & 31.25 \\
LSS~\cite{hahner2022lidar} & \underline{17.43} & \underline{36.77} & \underline{9.93} & 1.52 & \underline{41.03} & 22.74 & 30.82 \\
None & 16.29 & 34.82 & 8.05 & 1.44 & 35.98 & 20.30 & 28.84 \\
\rowcolor[gray]{0.9}
WeatherGen & \textbf{17.68} & \textbf{36.90} & \textbf{11.18} & \textbf{1.92} & \textbf{42.14} & \textbf{23.99} & \textbf{34.57} \\
\textit{Improved} & \textit{+1.39} & \textit{+2.08} & \textit{+3.13} & \textit{+0.48} & \textit{+6.15} & \textit{+3.69} & \textit{+5.73}
\\ \bottomrule
\vspace{-1.5cm}
\label{tab: dense_fog}
\end{tabular}}
\end{table}
\begin{table}[!t]
\huge
\centering
\renewcommand\arraystretch{1.15}
\caption{
Comparison with simulation-based methods on Seeing Through Fog~\cite{bijelic2020seeing} dataset under snow heavy test split.
}
\vspace{-0.1cm}
\resizebox{0.48\textwidth}{!}{
\begin{tabular}{ccccc|ccc}
\bottomrule
\multirow{2}{*}{Method}  & \multicolumn{7}{c}{Heavy snowfall scenarios$\uparrow$} \\ 
\cmidrule(r){2-8}
& 0-80m & 0-30m & 30-50m & 50-80m & \makecell[c]{Car AP\\ @.5IoU} & \makecell[c]{Cyc. AP\\ @.25IoU} & \makecell[c]{Ped. AP\\ @.25IoU} \\ \hline
DROR~\cite{charron2018noising} & 29.41 & \textbf{54.01} & 21.97 & 5.10 & 69.22 & 29.31 & 37.30 \\
LISA~\cite{kilic2021lidar} & 28.22 & 49.96 & 25.01 & 5.32 & 72.23 & \underline{36.28} & \underline{38.07} \\
FSRL~\cite{hahner2021fog} & \underline{30.03} & 51.92 & 23.95 & \textbf{5.54} & 71.62 & 33.23 & 37.38 \\
LSS~\cite{hahner2022lidar} & 29.97 & 52.24 & \underline{25.15} & 5.26 & \textbf{73.92} & 34.09 & 37.66 \\
None & 28.07 & 50.64 & 23.13 & 5.08 & 72.10 & 31.74 & 37.32 \\
\rowcolor[gray]{0.9}
WeatherGen & \textbf{30.98} & \underline{53.68} & \textbf{25.98} & \underline{5.33} & \underline{72.48} & \textbf{39.51} & \textbf{38.66} \\
\textit{Improved} & \textit{+2.91} & \textit{+3.04} & \textit{+2.85} & \textit{+0.25} & \textit{+0.38} & \textit{+7.77} & \textit{+1.34} 
\\ \bottomrule
\vspace{-1.5cm}
\label{tab: heavy_snow}
\end{tabular}}
\end{table}
\begin{table}[!t]
\Huge
\centering
\renewcommand\arraystretch{1.15}
\caption{
Comparison with simulation-based methods on Seeing Through Fog~\cite{bijelic2020seeing} dataset under rain test split.
}
\vspace{-0.1cm}
\resizebox{0.48\textwidth}{!}{
\begin{tabular}{ccccc|ccc}
\bottomrule
\multirow{2}{*}{Method}  & \multicolumn{7}{c}{Rain scenarios$\uparrow$} \\ 
\cmidrule(r){2-8}
& 0-80m & 0-30m & 30-50m & 50-80m & \makecell[c]{Car. AP\\ @.5IoU} & \makecell[c]{Cyc. AP\\ @.25IoU} & \makecell[c]{Ped. AP\\ @.25IoU} \\ \hline
LISA~\cite{kilic2021lidar} & \underline{34.65} & \underline{56.19} & \underline{25.21} & \underline{13.09} & 73.49 & \underline{34.27} & \underline{36.63}\\
None & 33.69 & 52.04 & 24.69 & 13.01 & \underline{74.02} & 34.02 & 35.01 \\
\rowcolor[gray]{0.9}
WeatherGen & \textbf{36.00} & \textbf{56.56} & \textbf{26.96} & \textbf{18.57} & \textbf{75.32} & \textbf{35.54} & \textbf{37.94} \\
\textit{Improved} & \textit{+2.31} & \textit{+4.52} & \textit{+2.27} & \textit{+5.56} & \textit{+1.30} & \textit{+1.52} & \textit{+2.93}
\\ \bottomrule
\vspace{-1.5cm}
\label{tab: rain_all}
\end{tabular}}
\end{table}

\subsection{Complexity Analysis}
\begin{figure}\hspace{-2.6mm}
\includegraphics[width=0.47\textwidth]{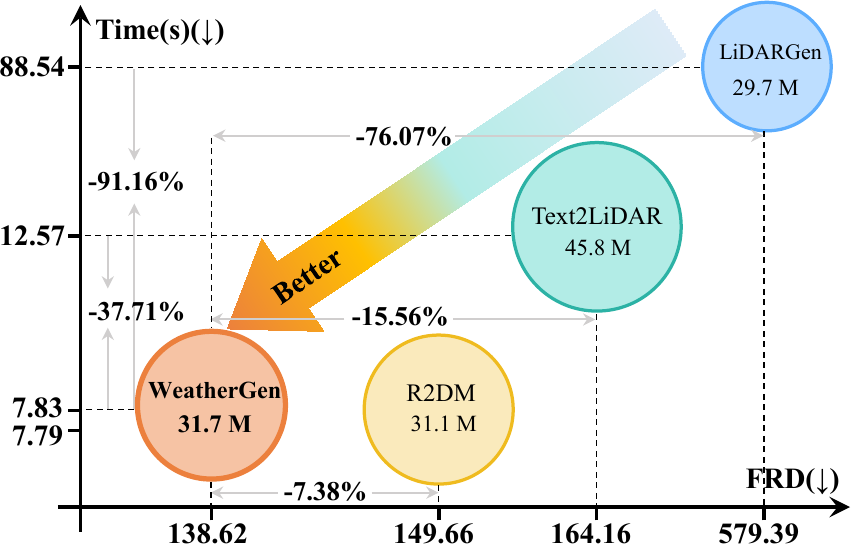}
\caption{Model efficiency comparisons. WeatherGen balances performance with a relatively small number of parameters.}
\vspace{-0.6cm}
\label{fig: para}
\end{figure}

Figure~\ref{fig: para} shows the comparison of model parameters and generation speed between our method and unconditional generation methods~\cite{nakashima2023lidar, zyrianov2022learning, wu2024text2lidar}.
%
%
%
Compared to the R2DM, our method achieved a $7.38\%$ performance improvement with only a $0.6$ M increase in parameters. 
Compared to LiDARGen, the first method in this field to employ diffusion technology, WeatherGen achieves comprehensive advantages. WeatherGen achieves a $76.07\%$ performance improvement and a $91.16\%$ enhancement in speed.
%
%
We also conduct an efficiency comparison for the Transformer-based Text2LiDAR. The results show that the Transformer architecture has disadvantages in terms of performance and generation speed.
The above illustrates that WeatherGen achieves the best speed, parameters, and performance balance, reaching a leading level at present. 


\section{Conclusion}
%
In this paper, we have proposed WeatherGen, the first unified framework for diverse weather LiDAR data generation. 
Within WeatherGen, an MDP has been designed to produce diverse-weather LiDAR data in a map-to-map manner, providing sufficient training data with high fidelity and diversity.
Among the diffusion network, an SMG is designed together with the spider mamba to restore the diverse weather data. By scanning LiDAR beams in a per-beam manner, SMG can maintain the physical structure of the LiDAR data excellently. Following that, an LFA is proposed to embed real-world knowledge into the model. 
Following that, we have designed a CLC to equip the control signal with compact semantic knowledge through contrastive learning. CLC enables the generator to generate higher discriminative LiDAR data.
Extensive experiments and the proposed mini-weather dataset have demonstrated the superiority of its generation quality and its immense value for downstream tasks under adverse weather conditions.

\textbf{Acknowledgments.} This work is supported by the National Key R\&D Program of China No. 2024YFC3015801, NSFC under Grant Nos. 62361166670, 62276144, 62276141, 62176124 and U24A20330.

{
    \small
    \bibliographystyle{ieeenat_fullname}
    \bibliography{main}
}


\end{document}